\pdfoutput=1
\documentclass[11pt]{article}

\usepackage[]{acl}

\usepackage{times}
\usepackage{latexsym}

\usepackage{booktabs}
\usepackage{footnote}

\usepackage{amsmath,amssymb,multicol,mathrsfs}
\usepackage[ruled,linesnumbered]{algorithm2e}
\usepackage{graphicx}
\usepackage{balance}

\usepackage{epstopdf}
\usepackage{multirow}
\usepackage{url}
\usepackage[skip=0pt]{subcaption}
\usepackage{color,soul}

\usepackage{tabularx}
 
\usepackage[normalem]{ulem}
\usepackage{enumitem}
\setlist{leftmargin=5.5mm}

\usepackage{url}

\usepackage[T1]{fontenc}
\usepackage[utf8]{inputenc}

\usepackage{microtype}

\title{Event Detection Explorer: \\ An Interactive Tool for Event Detection Exploration}

\author{Wenlong Zhang$^1$, Bhagyashree Ingale$^1$, Hamza Shabir$^1$, Tianyi Li$^2$, Tian Shi$^3$, Ping Wang$^1$\\
$^1$Dept. of Computer Science, Stevens Institute of Technology, Hoboken, NJ \\
$^2$Dept. of Computer and Information Technology, Purdue University, West Lafayette, IN\\
$^3$Moody's Analytics, New York, NY \\
{\tt \{wzhang71,bingale,hbuch,ping.wang\}@stevens.edu} \\
{\tt li4251@purdue.edu,researchtianshi@gmail.com} \\}

\begin{document}
\maketitle
\begin{abstract}
Event Detection (ED) is an important task in natural language processing. In the past few years, many datasets have been introduced for advancing ED machine learning models. However, most of these datasets are under-explored because not many tools are available for people to study events, trigger words, and event mention instances systematically and efficiently. In this paper, we present an interactive and easy-to-use tool, namely ED Explorer, for ED dataset and model exploration. ED Explorer consists of an interactive web application, an API, and an NLP toolkit, which can help both domain experts and non-experts to better understand the ED task. We use ED Explorer to analyze a recent proposed large-scale ED datasets (referred to as MAVEN), and discover several underlying problems, including sparsity, label bias, label imbalance, and debatable annotations, which provide us with directions to improve the MAVEN dataset.
The ED Explorer can be publicly accessed through \url{http://edx.leafnlp.org/}.
The demonstration video is available here \url{https://www.youtube.com/watch?v=6QPnxPwxg50}.
\end{abstract}

\section{Introduction}
Being one of the basic elements of event understanding, event detection (ED) aims at detecting event triggers from unstructured texts and classifying them into some predefined event types  \cite{chen-etal-2017-automatically,le2021fine}. 
It is one of the most important steps for extracting structured event information from unstructured texts \cite{ahn2006stages}.
% What about this:
% Event detection (ED) is one of the most important steps for extracting structured event information from unstructured text data \cite{ahn2006stages}. As one of the basic elements of event understanding, event detection extracts \textit{event triggers}, which are the word or phrase that indicates an event occurrence, and classifies them into predefined \textit{event types} \cite{chen-etal-2017-automatically,le2021fine}. 
Efficient and accurate event detection will also benefit many Natural Language Processing (NLP) tasks, such as information retrieval \cite{jungermann2008enhanced,kanhabua2016temporal}, question answering \cite{yang2003structured,souza2020event}, and event augment prediction \cite{cheng2018implicit}. 

\begin{figure}[!tp]
	\centering
	\includegraphics[width=0.9\linewidth]{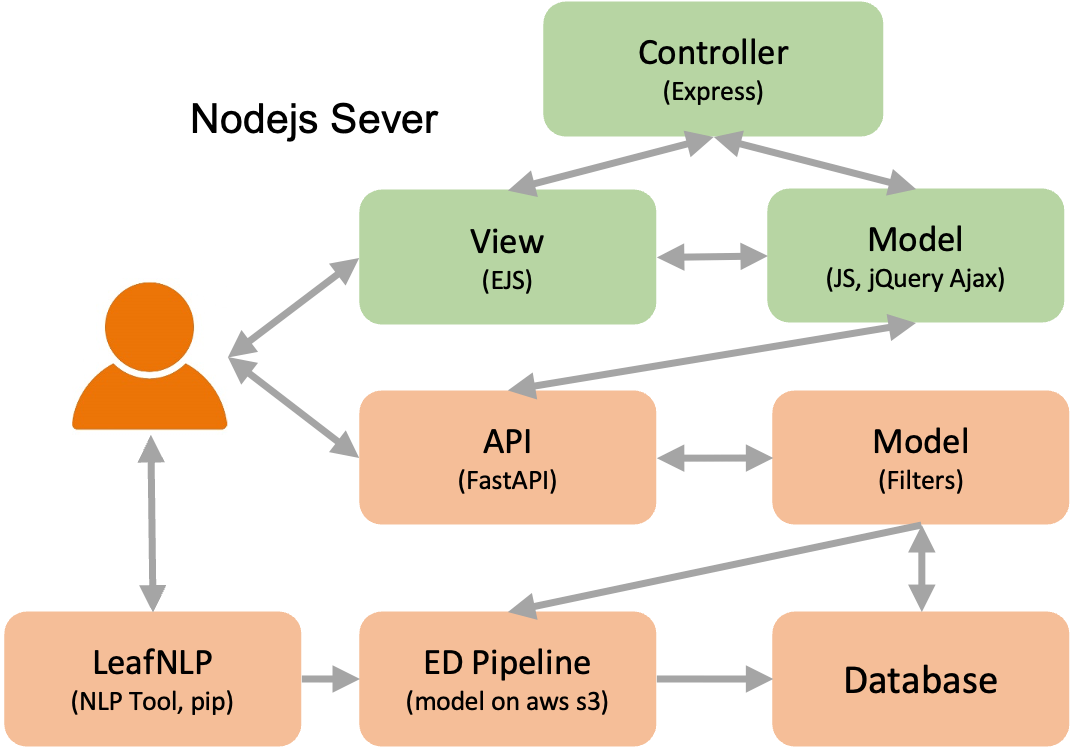}
% 	\vspace{-2mm}
	\caption{The architecture of the ED Explorer.}
	\label{fig:ed_architecture}
% 	\vspace{-4mm}
\end{figure}

The ED task has attracted considerable attention in recent years.
Traditional feature-based models \cite{araki2015joint, li2013joint, gupta2009predicting} rely on constructing different features that are related to events and incorporating them into the models.
Many recent deep learning based models formulate the ED task as a sequence labeling problem and have achieved state-of-the-art results \cite{lai2020event,liu2019exploiting,yan2019event,ding2019event,zhao2018document,chen2018collective}.
The advances of deep ED models are attributed to the development of datasets that can be used to train and benchmark these models.
In the past decades, several ED datasets have been introduced and widely used to develop ED models, such as {ACE 2005} \cite{walker2006ace} and {TAC KBP} \cite{mitamura2015overview}.

However, these datasets suffer from several limitations \cite{wang2020maven}.
(1) Data Scarcity.
These datasets are in small scale and cover a small number of instances. 
For example, there are only 599 documents and 5,349 instances in ACE 2005.
(2) Low Event Type Coverage, i.e.,
only a small number of event types are considered in these datasets. 
For example, ACE 2005 and TAC KBP have only 33 and 38 event types, respectively.
(3) Label Imbalance. 
In these datasets, many events are related to certain topics, which results in the label imbalance problem.
% What about ...
% The number of event mention instances for each event type differ dramatically in these datasets.
For example, in ACE 2005, $60\%$ event types have less than 100 annotated event mention instances.

Recently, a large scale ED dataset, namely MAVEN \cite{wang2020maven}, has been introduced.
It has more than 100K event mention instances for 168 event types, which
alleviates the Data Scarcity and Low Event Type Coverage problems.
There are also several other datasets that can be used to train and evaluate ED models \cite{sims-etal-2019-literary,lee2021effective,satyapanich2020casie}.
For example, RAMS \cite{ebner2020multi}) was originally annotated for document-level argument linking.
It has 9,124 annotated events across 139 types and can also be used to train ED models.
ALDG \cite{chen-etal-2017-automatically} and FewEvent \cite{deng2020meta} are automatically labeled datasets, which are used as augmented datasets to improve ED models.

Despite more ED datasets are made available for research and different models have been developed based on them \cite{yu2021lifelong,wang2021behind},
there are still a few problems that need to be investigated:
% The logic above reads a bit confusing to me, what about ...
% As more ED datasets are being developed and used to train different ED models~\cite{yu2021lifelong,wang2021behind}, it is important to be able to assess and compare the new datasets effectively and efficiently. Especially, the following aspects are important for understanding an ED dataset. 
(1) Uniqueness. What are the advantages of each ED dataset compared with other datasets (including ACE 2005 and TAC KBP) in terms of event types, trigger words, data distributions, event type coverage, and practical applications?
(2) Reliability. Since all these new ED datasets are recently introduced, they have not been comprehensively validated by other domain experts.
It is unclear if these datasets also suffer from data bias \cite{wang2021behind}, label imbalance, and annotation artifact \cite{gururangan2018annotation} problems.
(3) Accessibility.
Although there are many tools and packages for domain experts to explore and visualize these ED datasets, it is still difficult for most people to process and analyze them, and understand the ED task. 
% the sentence above didn't say *why* the existing tools are insufficient, do you mean ...
% While software packages have been developed for domain experts to explore and visualize ED datasets, it usually requires nontrivial installation and configuration efforts to use the tools. This makes the tools exclusive to experienced experts and hinders broader participation in dataset analysis. Furthermore, the existing tools are designed to explore one dataset at a time, and cannot be readily used to assess and compare multiple ED datasets. 
Therefore, can we develop a tool that can help them systematically explore ED datasets, so that this task can be easily accessed by more people?

\begin{figure*}[!tp]
	\centering
	\includegraphics[width=.9\linewidth]{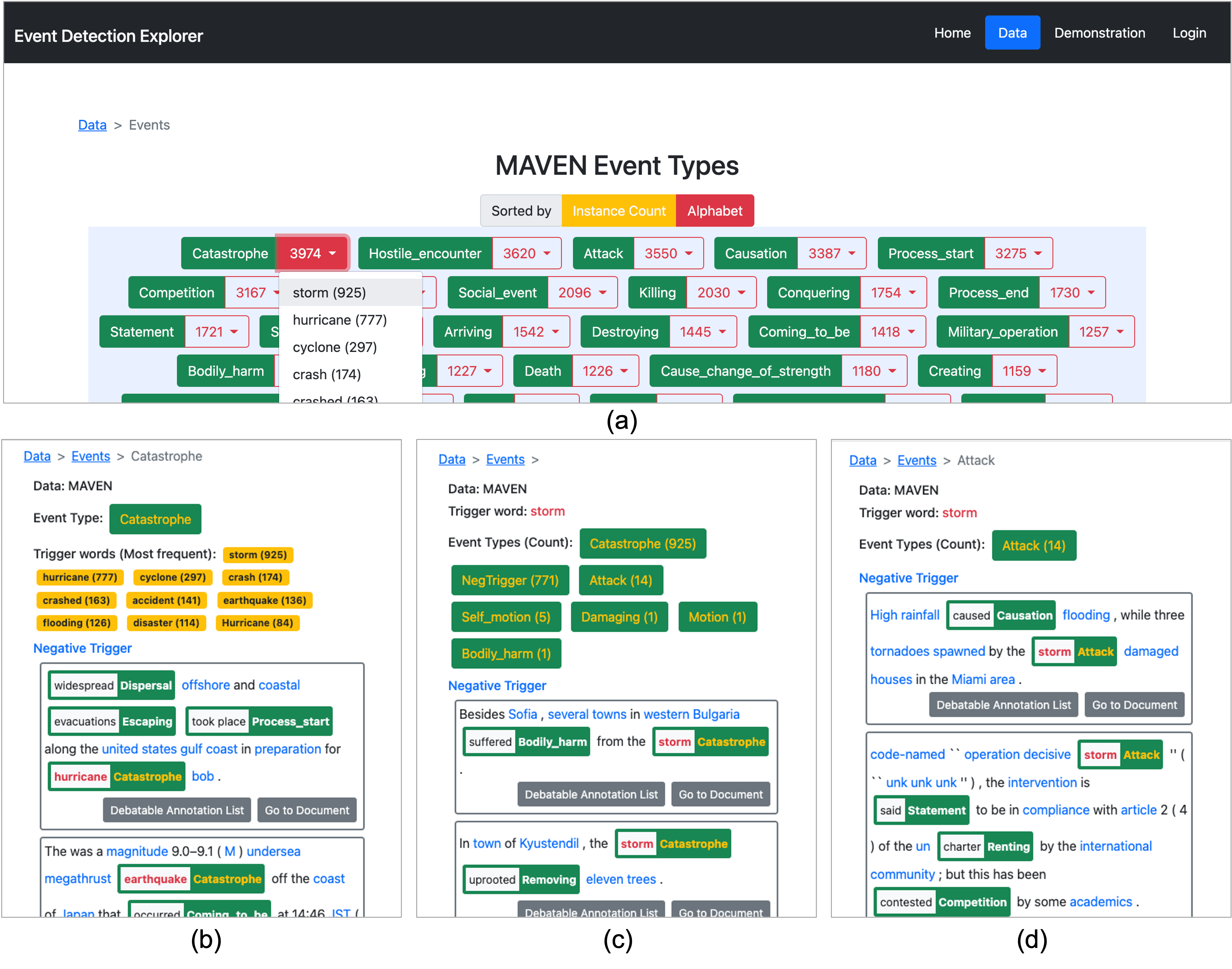}
	\vspace{-3mm}
	\caption{Front-end design of ED Explorer, which includes three primary components, including (a) Events Overview, (b) Event Type Explorer, and  (c-d)~Trigger Word Explorer.} 
	\label{fig:ed_front_overview}
% 	\vspace{-4mm}
\end{figure*}

To address these problems, we develop an ED Explorer (see Fig.~\ref{fig:ed_architecture}), that allows both domain experts and non-experts to systematically and efficiently explore different publicly available ED datasets, and the models trained on them.
There are three toolkits for users: A web application, an API, and a NLP toolkit.
The interactive front-end of the web application (see Fig.~\ref{fig:ed_front_overview}) makes it very easy for end users to navigate between different event types and trigger words, which can help them better understand the datasets and efficiently check and discover underlying problems in the annotations.
There is also a home maintained and easy-to-use NLP toolkit in Python, namely LeafNLP, for the ED task.
Therefore, users can test the ED models via the integrated and interactive web application (see Fig.~\ref{fig:ed_demo}), API and LeafNLP.

\section{Event Detection Datasets}
\label{sec:datasets}

In this section, we introduce the details of three representative ED datasets that are presented in our ED Explorer, including MAVEN \cite{wang2020maven}, RAMS \cite{ebner2020multi}, and ALDG \cite{chen-etal-2017-automatically}.
MAVEN represents open-domain general purpose ED datasets, which can detect multiple triggers and events in a single sentence.
For RAMS and ALDG, each of the sentences has only one primary event mention.
RAMS is manually annotated which is more reliable,
while ALDG is automatically generated and can be used in data augmentation when training ED models.
There are also several other publicly available datasets, such as CASIE \cite{satyapanich2020casie} and Commodity News Corpus \cite{lee2021effective}.
We will incorporate them to our platform in the future.
Since our platform is designed to be freely and publicly accessible, we do not include the well-known ED datasets ACE 2005 and TAC KBP. 
Table \ref{tab:data-stats} provides the basic statistics of these datasets.

\begin{table}[!t]
    \centering
    \resizebox{\linewidth}{!}{
    \begin{tabular}{l|ccc}
         \toprule
         \bf Datasets & \bf MAVEN & \bf RAMS & \bf ALDG \\\hline
         Domain & Wikipedia & News &Wikipedia   \\
        %  \# Documents &4,480 &3,993 &6.3M \\
         \# Sentences &49,873  &9,124 &72,611 \\
         \# Event types &168  &139 &21 \\
         \# Event mentions &118,732  &9,124 &72,611 \\
         \bottomrule
    \end{tabular}}
    \vspace{-2mm}
    \caption{Basic statistics of the three ED datasets used.}
    \label{tab:data-stats}
    \vspace{-4mm}
\end{table}

\begin{itemize}[leftmargin=*,topsep=0pt,itemsep=1pt,partopsep=1pt, parsep=1pt]
    \item \textbf{MAssive eVENt detection (MAVEN) dataset} \cite{wang2020maven}: is a massive ED dataset developed in 2020 by combining machine generation and human-annotation based on 4,480 Wikipedia documents.
    It aims at addressing limitations of existing ED datasets about data scarcity and low coverage of event types. The  event types in MAVEN are derived from the frames defined in the linguistic resource Frame net \cite{baker1998berkeley} with a large coverage of events in the general domain.
    Compared with existing datasets, MAVEN covers 168 event types, and 118,732 events mentions, which indicates a larger data scale and a larger event coverage. Recently, it has been used for developing different ED models \cite{cao2021knowledge,yu2021lifelong,wang2021behind,frisoni2021survey,wang2021behind}. 

    \item \textbf{Roles Across Multiple Sentences (RAMS)} \cite{ebner2020multi}: is a crowdsourced dataset developed for identifying explicit arguments of different roles for an event from multiple sentences, which is known as multi-sentence argument linking. It covers 139 event types, 9,124 annotated events from 3,993 news articles, and 65 roles. Compared with prior small-scale datasets for cross-sentence argument linking, RAMS advances the development of advanced deep learning models for this task \cite{zhang2020two,wen2021resin,lou2021mlbinet}.

    \item \textbf{Automatically Labeled Data Generation for large scale event extraction (ALDG)} \cite{chen-etal-2017-automatically}: is an automatically generated dataset using distant supervision \cite{mintz2009distant} by jointly using the world knowledge Freebase \cite{bollacker2008freebase} and linguistic knowledge FrameNet. ALDG covers 72,611 events across 6.3 million articles in Wikipedia and 21 event types with a focus on the topic about education, military, and sports.
    
\end{itemize}

\section{ED Explorer}
\label{sec:system}
In this section, we describe the pipelines and usage of our Event Detection (ED) Explorer.

\subsection{Architecture}
Our ED Explorer (see Fig.~\ref{fig:ed_architecture}) enables end-users to explore ED datasets and models by interacting with a Web Application and API.
The Web Application is an HTTP server in Node.js developed following the concept of Model-View-Controller design pattern.
For the Controller, we adopt \textit{express} as the primary framework and \textit{express router} to handle routing and user navigation.
For the View (i.e., front-end), we use \textit{ejs} as our template engine to generate HTML and use Bootstrap to style web pages.
For the Model, different models do not directly interact with databases, instead they send and receive JSON content via HTTP requests (e.g., GET and POST) to a REST API.
The Web API is built with FastAPI, which is a high-performance Python web framework.
Different models, i.e., functions, in FastAPI handle different requests, interact with databases or machine learning (ML) pipelines, and respond to the requests.
The ML pipelines (1) get messages from API models, (2) annotate texts with a home maintained NLP toolkit, namely LeafNLP\footnote{\url{https://pypi.org/project/leafnlp}}, and (3) store the annotated results in a database.

Following this design, there are three entry points for end-users: (1) Web Application. Users can explore ED datasets and models via front-end of ED Explorer.  (2) Web API. Users can get processed data and output of ED models via interacting with the Web API. (3) LeafNLP. Users can download and install LeafNLP (via \textit{pip install leafnlp}) and use this toolkit to annotate their text.

\subsection{ED Dataset Explorer}

The ED Dataset Explorer (EDDE, see Fig.~\ref{fig:ed_front_overview}) aims at helping users understand and explore ED datasets more efficiently.
In this section, we introduce three primary components of EDDE.

\subsubsection{Events Overview}
The first component of EDDE is an events overview page (see Fig.~\ref{fig:ed_front_overview}~(a)) that shows the distribution of different events, including distributions of trigger words in different events. The design of this page is intended to help end-users understand the following research questions:
\begin{itemize}[leftmargin=*,topsep=0pt,itemsep=1pt,partopsep=1pt, parsep=1pt]
    \item How many event types are there? What are they?
    \item How many event mention instances for each event type?
    \item What are trigger words for each event type?
    \item How many event mention instances for each event type and trigger word?
\end{itemize}
On one hand, these questions are very important for us to find underlying problems of an ED dataset and improve existing annotations.
On the other hand, they provide us some guidance to develop and evaluate ML models.
For example, in MAVEN dataset, more than 30 event types, such as \textit{Breathing} and \textit{Change Tool}, have less than 100 event mention instances.
Therefore, we should perform extra evaluations for ML models in predicting these rare event types.
Also, we should annotate more instances for these events in order to improve the quality of the MAVEN dataset.

\subsubsection{Event Type Explorer}

Fig.~\ref{fig:ed_front_overview}~(b) displays the second component of EDDE, namely, Event Type Explorer.
For each event type (e.g., \textit{Catastrophe}), we show 10 most frequent trigger words (with the count of their corresponding instances) and all event mention instances (i.e., sentences).
In each instance, we use RED color to highlight trigger words, and also show other triggered events and their trigger words.
In addition, we use BLUE color to indicate candidate trigger words that do not trigger any event (namely, negative triggers).
With Event Type Explorer, end-users can easily and efficiently explore different event types, their trigger words and event mention instances.

\subsubsection{Trigger Word Explorer}

\begin{figure}[!tp]
	\centering
	\includegraphics[width=0.9\linewidth]{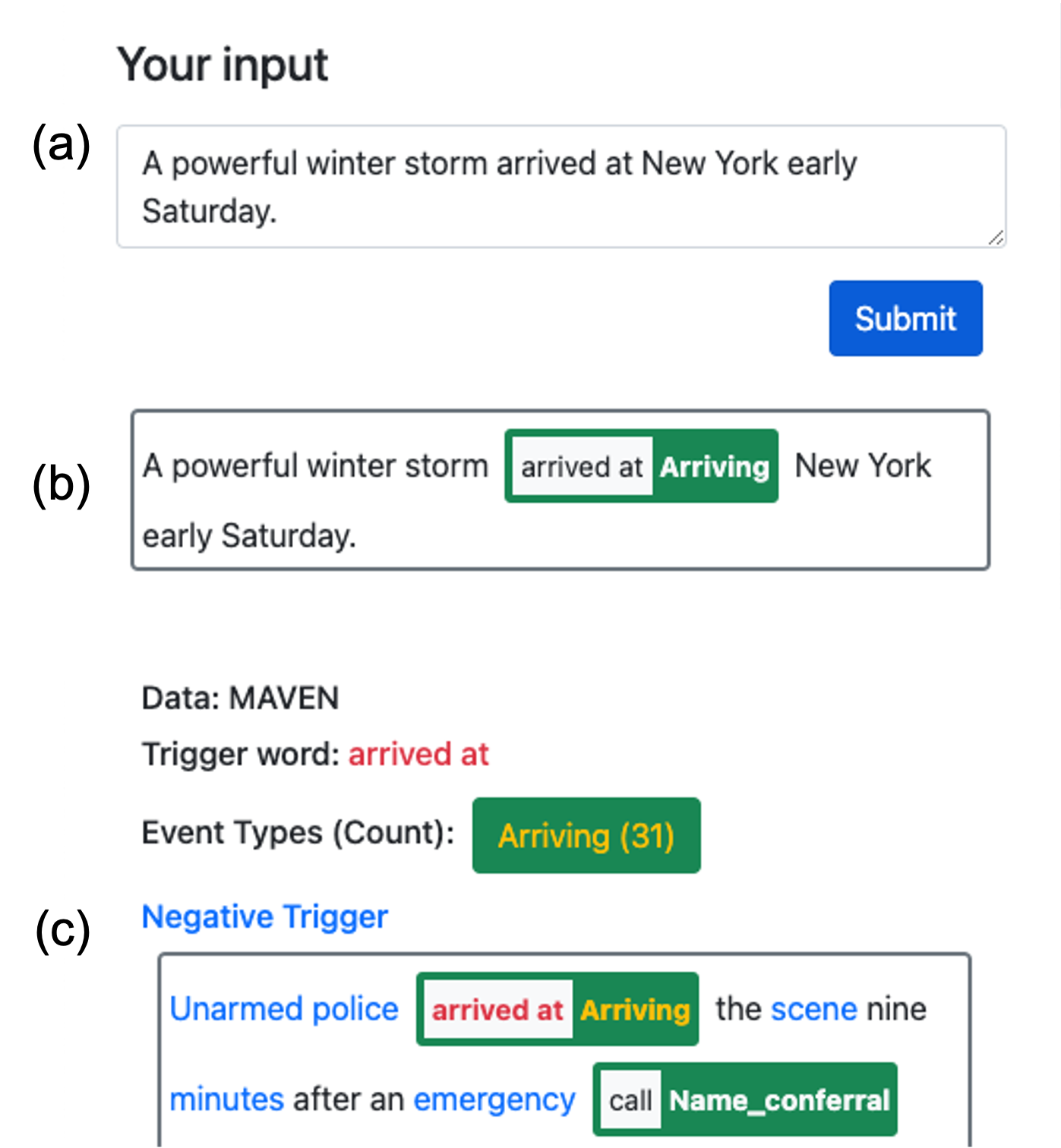}
	\caption{Live demonstration of ED Model Explorer. (a)~Input sentence or article. (b) Output from ED Model Explorer with predicted events. (c) Further exploration of the trigger word with Trigger Word Explorer.}
	\label{fig:ed_demo}
% 	\vspace{-4mm}
\end{figure}

Trigger Word Explorer, which is the third component of EDDE (Fig.~\ref{fig:ed_front_overview}~(c) and (d)), is designed to systematically explore trigger words, their event types, and event mention instances.
In most real-world applications (e.g., Fig.~\ref{fig:ed_demo}~(b)), we start exploring events by analyzing trigger words.
Therefore, understanding what events a word may trigger is naturally the first step to understand event detection.
The design of instance visualization is the same as that for Event Type Explorer. 
Fig.~\ref{fig:ed_front_overview}~(c) and (d) represent two different filters, which show event mention instances for all events triggered by a trigger word (i.e., \textit{storm}) and a single event (\textit{Attack}) triggered by the trigger word (\textit{storm}).

Trigger Words Explorer has played a very important role for us to identify incorrect annotations in this work.
For example, in Fig.~\ref{fig:ed_front_overview}~(c), we find that in 925 instances, \textit{storm} is annotated as \textit{Catastrophe}, but it is also labeled as \textit{Attack}, \textit{Self motion}, \textit{Damaging}, \textit{Motion}, and \textit{Bodily Harm}.
By manually checking these rare instances, we have found many incorrect annotations.
In 771 instances, \textit{storm} is treated as negative trigger, which is also questionable.

\subsection{ED Model Explorer}

In addition to ED Dataset Explorer, it is also important to explore machine learning (ML) models trained on an ED dataset, so that we can perform integrated analysis on model outputs, and jointly improve both data annotations and ML models.

In this work, we have implemented a deep learning model for the event detection task and trained it on the MAVEN dataset.
The architecture of the model is similar to the one used for the named entity recognition task in \cite{devlin2019bert}, where input texts are first tokenized with a case-preserving Word-Piece model; and then, they are encoded by a BERT encoder with 12 transformer layers \cite{wolf-etal-2020-transformers}.
The contextual embeddings from BERT encoder are passed to a randomly initialized two-layer BiLSTM before the classification layer.

Fig.~\ref{fig:ed_demo}~(a) and (b) show the front-end of the demonstration, where end-users can type their input (e.g., a sentence or an article) in the text box and then they can view the annotated sentences.
The trigger words and event types in Fig.~\ref{fig:ed_demo}~(b) has linked to the Trigger Words Explorer and Event Types Explorer, which makes it easy for them to check the dataset that the ED model was trained on (Fig.~\ref{fig:ed_demo}~(c)).
Therefore, this integrated interactive system can help end-users better understand the model outputs and the ED datasets.
\section{ED Datasets Analysis}
\label{sec:experiment}

With our ED Explorer, we have systematically explored MAVEN, RAMS, and ALDG datasets, and placed special emphasis on the MAVEN dataset, since it is developed and manually annotated for the ED task.
In this section, we present our primary findings for MAVEN.

\subsection{Common Problems}
\label{sec:common_problem}

\begin{table}[!tp]
    \centering
    \resizebox{\linewidth}{!}{
    \begin{tabular}{m{4em}|m{13em}|c}
    \toprule
        \bf Triggers & \bf Event Types (\# Ins.) & \bf N.T. (\# Ins.)  \\\hline
        
        crash &
        Catastrophe(174),
        Damaging(4)
        
        Motion(2), 
        Attack(2) & 153\\\hline
        
        damage &
        Damaging(619),
        Causation(1) 
        
        Destroying(1),
        Bodily Harm(1)  & 275\\\hline
  
        storm & 
        Catastrophe(925),
        Attack(14)
        
        Self Motion(5),
        Damaging(1)
        
        Motion(1),
        Bodily Harm(1)& 771 \\
        \bottomrule
    \end{tabular}}
    % \vspace{-2mm}
    \caption{Examples of trigger words with their annotated event types and frequencies in MAVEN. \textbf{N.T.} represents Negative Triggers that are note annotated.}
    \label{tab:data_trigger_events}
    \vspace{-4mm}
\end{table}

Through Event Overview page and Trigger Word Explorer, we found that it is necessary to investigate the statistics of the MAVEN dataset, because we observed that many event types only have less than 100 annotated instances; and for many trigger words, they trigger one event in most instances.
For example, \textit{crash} trigger \textit{Catastrophe} in 174 instances and is \textit{Negative Trigger} in 153 instances (see Table~\ref{tab:data_trigger_events}).
Here, we summarize the common problems as follows:

\begin{itemize}[leftmargin=*,topsep=0pt,itemsep=1pt,partopsep=1pt, parsep=1pt]
    \item \textbf{Sparsity}.
    In the MAVEN training set, there are 50,388 unique candidate trigger words, out of which 7,074 words trigger at least one event.
    The total number of annotated instances is 96,897.
    Within 7,074, only 963 words (14\%) have at least 20 annotated instances.
    However, they cover 75,950 annotated instances (78\%).
    Therefore, for most trigger words, they have very few instances to train ED models.
    \item \textbf{Label Bias}.
    In our ED Explorer, we also show the distribution of topics for documents used by MAVEN.
    We observed that most documents are about \textit{military conflict}, \textit{hurricane}, \textit{civilian attack}, \textit{concert tour} and \textit{civil conflict}, which may lead to label bias problem and limit the applications of ED models trained on MAVEN.
    For example, for event \textit{Building}, the most common triggers are \textit{established}, \textit{built}, \textit{building}, \textit{constructed}, \textit{build}, \textit{establish}, \textit{buildings}, \textit{set up}, \textit{assembly}, and \textit{erected}.
    In our experiments, we found via our ED Model Explorer that the ED model cannot detect event \textit{Building} in any of these sentences: ``We will build a house.'',
    ``We will construct a new building.'',
    ``We will expand the runway.'',
    ``We will redevelop the terminal.''
    \item \textbf{Label Imbalance}.
    For 7,074 words, we further checked the events they may trigger and found that 4,648 words (66\%) have triggered only one event in different instances.
    For other words that trigger more than one event, many of them have \textbf{dominant events}.
    Here, we define an event as the dominant event for a trigger word, if the number of instances for one event is significantly larger than other events (i.e., $\frac{\#\text{of instances for the most frequent events}}{\# \text{of instances for other events}}>5$).
    For example, in Table~\ref{tab:data_trigger_events}, \textit{Catastrophe} is the dominant events for \textit{crash} and \textit{storm}.
    Using this definition, we find that among the 963 words that have more annotated instances, 585  (61\%) of them have dominant events, which results in the label imbalance problem.
    Therefore, the ED model trained on MAVEN may suffer from the problem that it predicts only one event for each trigger word in different scenarios.
\end{itemize}

\begin{table}[!tp]
    \centering
    \resizebox{\linewidth}{!}{
    \begin{tabular}{m{4.7em}|m{18em}}
    \toprule
        \bf Problems & \bf Instance Examples \\\hline
        
        \bf Negative Trigger & 
        It formed on October 1 in the Caribbean Sea as the seventeenth tropical \underline{\bf storm::Negative Trigger}, and initially moved slowly to the north. \\\hline
        
        \bf Trigger Wrong Events &
        Unknown to the hijackers, passengers aboard \underline{\bf made::Manufacturing} telephone calls to friends and family and relayed information on the hijacking. \\\hline
        
        \bf Events Ambiguity & 
        \textbf{S1:} The hurricane reached peak winds of 125 mph (205 km/h) on October 6 while \underline{\bf moving::Motion} through the Bahamas.
        
        \textbf{S2:} By midday on June 25, the hurricane reached peak winds of before \underline{\bf moving::Self Motion} inland well south of the U.S. Mexico border.\\
        
        \bottomrule
    \end{tabular}}
    % \vspace{-2mm}
    \caption{Common annotation problems in MAVEN. The pattern \textbf{moving::Motion} represents the word \textit{moving} triggers an event \textit{Motion}.}
    \label{tab:annotation_debatable}
    \vspace{-3mm}
\end{table}

\subsection{Debatable Annotations in MAVEN}

In addition to analyzing the distributions of events and triggers, we have also manually checked annotated instances in MAVEN.
In details, for each of the 168 event types, we chose the most frequent 10 trigger words as candidates and for each trigger word, we manually checked 10 annotated instances for every event type it triggers.
These procedures can be easily accomplished by navigating through ED Event Type (see Fig.~\ref{fig:ed_front_overview}~(b)) and Trigger Word Explorers (see Fig.~\ref{fig:ed_front_overview}~(c) and (d)).

We have checked around 10,000 instances in total, and found 2,579 debatable instances (25\%), which are also shown in the ED Explorer and are publicly available via API.
From these debatable annotations, we found that there are typically three types of annotation mistakes:
(1) \textbf{Negative Trigger} represents the situations that annotating a word triggers an event as a negative trigger.
(2) \textbf{Trigger Wrong Events} indicates that the word does not trigger the annotated event types.
(3) \textbf{Events Ambiguity} means that it is difficult to distinguish two event types (such as \textit{Motion} and \textit{Self Motion}), so the annotated instances are also debatable.
We have shown examples of each of the annotation problems in Table~\ref{tab:annotation_debatable}.

\subsection{Findings from the ED Model Explorer}

We have also found several annotation problems via ED Model Explorer, i.e., the ED Demonstration.
(1) The first problem is \textbf{Label Bias}, which has been discussed in Section~\ref{sec:common_problem}.
(2) The second problem is that many words that should trigger an event do not actually trigger any event in a number of testing cases.
For example, in Fig.~\ref{fig:ed_demo}, the word \textit{storm} should trigger an event \textit{Catastrophe}, however, it does not trigger any events, even if we test cases like ``The storm hits New York.'', ``The storm damages a lot of houses.'', etc.
We think this might because \textit{storm} is labeled as \textit{Negative Trigger} in 771 instances (see Table~\ref{tab:data_trigger_events}).

\subsection{Future Directions}

As ED Explorer can help us systematically explore ED datasets, better understand events and trigger words, and efficiently discover underlying annotation problems, we will include more ED datasets and models, and make it easier for people to explore them in a single integrated platform in the future. 
More importantly, we plan to improve the quality of MAVEN dataset by 
(1) continuing checking and correcting annotation errors, 
(2) annotating more documents in other fields to mitigate label bias problem,
(3) annotating more instances for non-dominant events to address label imbalance problem.

\section{Conclusion}
\label{sec:conclusion}

In this paper, we introduced an event detection exploration tool, namely ED Explorer, to help domain experts and non-experts better understand the ED task, ED datasets, and ED models.
ED Explorer consists of an interactive web application, an API, and a LeafNLP toolkit, which allows end users to access datasets and models in a variety of ways.
With ED Explorer, we conduct a detailed and systematic analysis on a recently developed MAVEN dataset and discover several underlying problems during annotations, including sparsity, label bias and imbalance, and annotations errors.
In the future, we will further improve the ED Explorer and address annotation problems in the MAVEN dataset.

\bibliography{7-ref}
\bibliographystyle{acl_natbib}

% \appendix
% \section{Example Appendix}
% \label{sec:appendix}
% This is an appendix.

\end{document}